\documentclass[a4paper, 10pt, conference]{ieeeconf}      % Use this line for a4 paper

\IEEEoverridecommandlockouts                              % This command is only needed if 
\usepackage{graphics} % for pdf, bitmapped graphics files
\usepackage{epsfig} % for postscript graphics files
\usepackage{mathptmx} % assumes new font selection scheme installed
\usepackage{times} % assumes new font selection scheme installed
\usepackage{amsmath} % assumes amsmath package installed
\usepackage{amssymb}  % assumes amsmath package installed
\usepackage{soul}
\usepackage{lipsum}
\usepackage[binary-units=true,per-mode=symbol]{siunitx}
\usepackage{epstopdf}
\usepackage[space]{cite}

\usepackage[nolist]{acronym} % manually added by Simon Hoffmann -> usage: https://www.namsu.de/Extra/pakete/Acronym.html
\usepackage{hyperref} % manually added by Simon Hoffmann -> use \autoref{}

\usepackage[colorinlistoftodos, textsize=small, textwidth=50mm]{todonotes}

\definecolor{ascolor}{rgb}{.7, 0.3, 0.0}

\definecolor{tofcolor}{rgb}{.1, 0.7, 0.1}

\title{\LARGE \bf
Open Source Software for Teleoperated Driving
}

\author{Andreas Schimpe, Johannes Feiler, Simon Hoffmann, Domagoj Majstorovi\'c and Frank Diermeyer% <-this % stops a space
\thanks{The authors are with the Institute for Automotive Technology at the Technical University of Munich (TUM), 85748 Garching bei M\"unchen, Germany. {\tt\small \{firstname\}.\{lastname\}@tum.de}}%
}

\begin{document}
\maketitle
\thispagestyle{empty}
\pagestyle{empty}

\begin{acronym}
%\acro{cc}[CC]{Control Center}
%\acroplural{cc}[CCs]{Control Centers}
%
%\acro{cr}[CR]{Control Room}
%\acroplural{cr}[CR]{Control Rooms}
%
%\acro{ddt}[DDT]{Dynamic Driving Tasks}
%\acroplural{ddt}[DDTs]{Dynamic Driving Tasks}
%\acro{oedr}[OEDR]{Object and Event Detection and Response}

\acro{av}[AV]{automated vehicle}
\acroplural{av}[AVs]{automated vehicles}

%\acro{avf}[AVF]{Automated Vehicle Fleet}
%\acroplural{avf}[AVF]{Automated Vehicle Fleets}

\acro{ad}[AD]{automated driving}

\acro{ads}[ADS]{automated driving system}
\acroplural{ads}[ADS]{automated driving systems}
\acro{ugv}[UGV]{unmanned ground vehicle}
\acroplural{ugv}[UGVs]{unmanned ground vehicles}

%\acro{odd}[ODD]{Operational Design Domain}
%
%\acro{rd}[RD]{Remote Driver}
%\acroplural{rd}[RD]{Remote Drivers}
%
%\acro{soc}[SOC]{State of Charge}
%
%\acro{ca}[CA]{Control Action}
%\acroplural{ca}[CAs]{Control Actions}
%
%\acro{cf}[CF]{Causal Factor}
%\acroplural{cf}[CFs]{Causal Factors}
%
%\acro{uca}[UCA]{Unsafe Control Action}
%\acroplural{uca}[UCAs]{Unsafe Control Actions}
%
%\acro{stpa}[STPA]{System-Theoretic Process Analysis}

\acro{hmd}[HMD]{head-mounted display}
\acroplural{hmd}[HMDs]{head-mounted displays}

%\acro{oem}[OEM]{Original Equipment Manufacturer}
%\acroplural{oem}[OEMs]{Original Equipment Manufacturer}
%
%\acro{cbw}[CbW]{Conduct by Wire}
%
%\acro{adas}[ADAS]{Advanced Driver Assistance Systems}

\acro{hmi}[HMI]{human-machine interface}

%\acro{oem}[OEM]{Original Equipment Manufacturer}
%\acro{ul}[UL]{up link}
%\acro{dl}[DL]{down link}

\acro{orccad}[ORCCAD]{Open Robot Controller Computer-Aided Design}

\acro{swa}[SWA]{steering wheel angle}
%\acro{sc}[SC]{System-level Constraint}
%\acroplural{sc}[SCs]{System-level Constraints}

%\acro{hazop}[HAZOP]{Hazard and Operability Study}
%\acro{fta}[FTA]{Fault Tree Analysis}
%\acro{fmea}[FMEA]{Failure Mode and Effect Analysis}
%\acro{stamp}[STAMP]{Systems-Theoretic Accident Model and Processes}
%
%%Actuators
%\acro{swOp}[SW]{steering wheel actuator}
%\acro{bpOp}[BP]{brake pedal actuator}
%\acro{tpOp}[TP]{throttle pedal actuator}
%\acro{imu}[IMU]{Inertial Measurement Unit}

%\acro{udp}[UDP]{User Datagram Protocol}
\acro{ecs}[ECS]{Entity Component System}
%\acro{usb}[USB]{Universal Serial Bus}
\acro{ros}[ROS]{Robot Operating System}
\acro{rwa}[RWA]{road wheel angle}

%\acro{tcp}[TCP]{Transmission Control Protocol}

\acro{mqtt}[MQTT]{Message Queuing Telemetry Transport}

%% Control Actions
%\acro{swaOp}[$\dot{\delta}_{\text{H,O}}$]{change steering wheel command}
%\acro{sBp}[$s_{\text{Bp}}$]{brake pedal travel}
%\acro{sTp}[$s_{\text{Tp}}$]{throttle pedal travel}
%\acro{vD}[$v_{\text{d}}$]{desired velocity}
%\acro{vA}[$v_{\text{a}}$]{actual velocity}
%\acro{swaD}[$\delta_{\text{d}}$]{desired steering wheel angle}
%\acro{swaA}[$\delta_{\text{a}}$]{actual steering wheel angle}
%\acro{swaM}[$M_{\text{H}}$]{steering torque}
%\acro{engineM}[$M_{\text{E}}$]{engine torque}

\acro{tod}[ToD]{Teleoperated driving}
%\acro{rtsp}[RTSP]{Real-Time Streaming Protocol}
%\acro{rtp}[RTP]{Real-Time Transport Protocol}
\acro{gui}[GUI]{graphical user interface}
\acro{g2g}[G2G]{glass-to-glass}

\end{acronym}
\begin{abstract}
Teleoperation allows a human operator to remotely interact with and control a mobile robot in a dangerous or inaccessible area. Besides well-known applications such as space exploration or search and rescue operations, the application of teleoperation in the area of automated driving, i.e., teleoperated driving~(ToD), is becoming more popular. Instead of an in-vehicle human fallback driver, a remote operator can connect to the vehicle using cellular networks and resolve situations that are beyond the automated vehicle~(AV)'s operational design domain. Teleoperation of AVs, and unmanned ground vehicles in general, introduces different problems, which are the focus of ongoing research. This paper presents an open source ToD software stack, which was developed for the purpose of carrying out this research. As shown in three demonstrations, the software stack can be deployed with minor overheads to control various vehicle systems remotely.
\end{abstract}

\section{\uppercase{Introduction}}
\label{sec:introduction}
\ac{tod} allows a human operator to remotely control an \ac{ugv}. In the context of \ac{ad}, \ac{tod} is a promising solution. To some extent, it can be used to replace in-vehicle fallback drivers because edge cases of the \ac{ad} function can be remotely resolved. \\
Whereas \ac{tod} has numerous possibilities, it also introduces new challenges, which are the focus of current research activities. These include vehicle control subject to latency conditions~\cite{Davis2010}, connection loss~\cite{Tang2014e}, management of limited network bandwidth~\cite{Schimpe2021}, or reduced operator situation awareness~\cite{Mutzenich2021}. The primary objective of this paper is the presentation and provision of a \ac{tod} software stack to the research community. Its usage is not inherently limited to \acp{av}, but may be extended to any type of on-road or off-road \ac{ugv}. Therefore, this paper presents software for carrying out research into various teleoperation concepts and vehicle platforms. 
\subsection{Related Work}
\label{sec:relatedwork}
Over the years, various teleoperation systems for \acp{ugv} have been presented. Bensoussan and Parent~\cite{Bensoussan1997} proposed a teleoperation setup for small, urban carsharing vehicles as long ago as 1997. The paper focuses on the hardware setup of the system. 
Gnatzig et al.~\cite{Gnatzig2013a} describe a system design for teleoperated road vehicles. In addition to hardware components, the work also presents and discusses the communication setup with the experienced bandwidth and latency of the mobile network. 
Bodell and Gulliksson~\cite{Bodell2016} describe simulation of the remote control of a truck. Different input devices were implemented to investigate their influence on teleoperation. The operator is provided with a stitched video stream containing additional information such as velocity or a map.
Shen et al.~\cite{Shen2016} describe a software and hardware architecture for teleoperating road vehicles in wireless networks. The video feeds of an actuated stereoscopic camera are visualized to the operator through a \ac{hmd}.
Georg and Diermeyer~\cite{Georg2019b} propose an immersive interface in a three-dimensional environment that provides the operator with information from various sensors. The interface, presented to the operator through multiple displays or an \ac{hmd}, can be adapted through scenes with different settings. 
At the time of publication of this paper, TELECARLA from Hofbauer et al.~\cite{Hofbauer2020} is the only available, open source software for \ac{tod}. Based on the \ac{ros}, it is an extension of the CARLA Simulator~\cite{Dosovitskiy2017}, providing the operator with an interface to directly steer and control the velocity of the vehicle. \\
The software stack, presented in this paper, is also based on \ac{ros} and was primarily developed for carrying out \ac{tod} research. However, interfaces between components follow established concepts of \ac{ad} functions, e.g., object lists or trajectories. This enables integration or extension of \ac{ros}-based \ac{ad} software such as the open source stacks Autoware~\cite{Kato2018} and Apollo~\cite{Baidu}.
\subsection{Contributions}
In this paper, a \ac{ros}-based software stack is presented the purpose of which is to support research in the field of \ac{tod}. The system design is modular, allowing easy integration with existing \ac{ad} software. With a conventionally designed vehicle interface, the software can be deployed with minor overheads to remotely control various vehicle systems. This is shown in three demonstrations. The source code is open source and available on GitHub\footnote{\href{https://github.com/TUMFTM/teleoperated\_driving}{https://github.com/TUMFTM/teleoperated\_driving}}.

\section{\uppercase{Robot Operating System}}
\ac{ros} is an open source software framework for scalable robot applications~\cite{Quigley2009}. It is well-established in academia, and is also becoming increasingly popular in industry. Use of \ac{ros} leverages the publisher-subscriber paradigm. Packages containing nodes can be seamlessly integrated into software that becomes complex, but still remains modular. With \ac{ros} providing a structured communication layer above the host operating system, the user/developer can focus more on the functionality of nodes and less on their interaction. Besides this, the \ac{ros} ecosystem also offers a wide range of tools, e.g., for data visualization and logging. 

\section{\uppercase{System Design}}
\label{sec:systemDesign}
\begin{figure}[!t]
    \includegraphics[width=\linewidth]{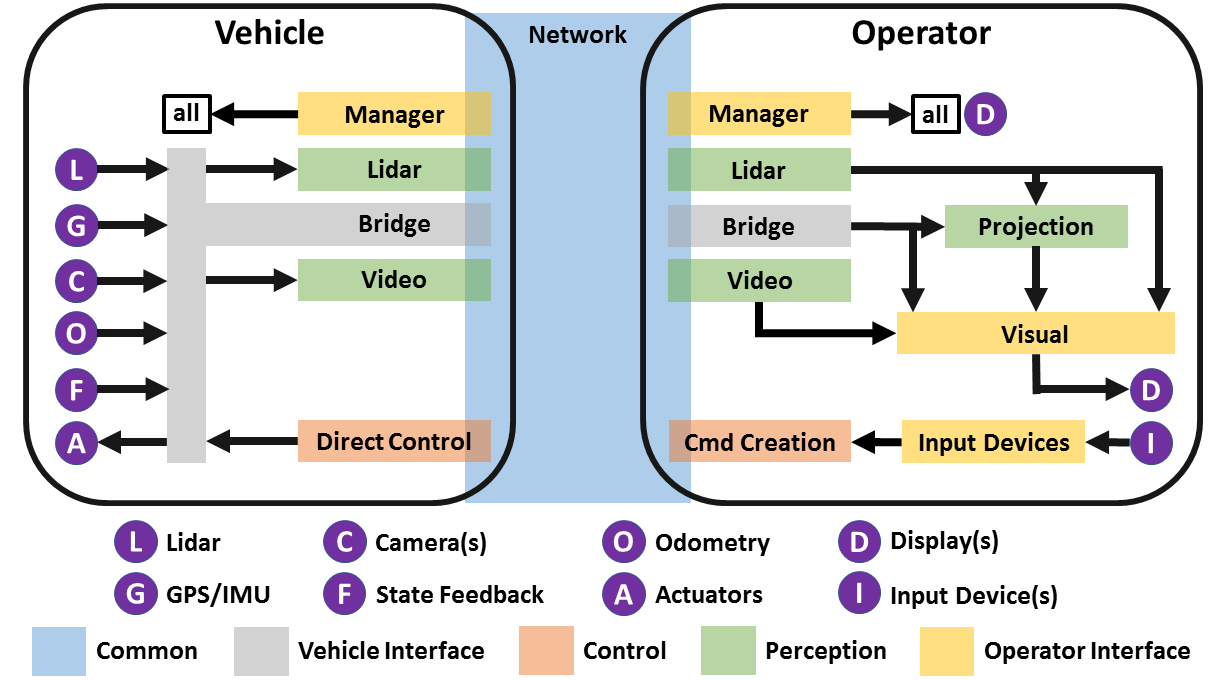}
    \caption{System architecture of the presented software. The purple circles represent data feeds from and to hardware components. The software packages, depicted as rectangles, are grouped and color-coded according to their high-level function.}
    \label{fig:architecture}
\end{figure}%
The complete system architecture is shown in Fig.~\ref{fig:architecture}. It is separated into vehicle and operator sides which are connected via the \textit{Network}~(blue). On the vehicle side, the \textit{Bridge} forms the \textit{Vehicle Interface}~(grey), interfacing various hardware components such as sensors and actuators. The \textit{Lidar}, \textit{Video} and \textit{Projection} packages form the \textit{Perception}~(green), responsible for processing the sensor data and transmitting it to the operator. The \textit{Operator Interface}~(yellow) comprises three packages. These are \textit{(1)}~the \textit{Manager}, connected to all other packages and managing the \ac{tod} session, \textit{(2)}~the \textit{Visual} package, that displays the output from the perception packages and other data from the vehicle, and \textit{(3)}~the \textit{Input Devices} that form the interfaces to various input device hardware components. From the output of the latter, the \textit{Control} packages~(orange), i.e., \textit{Command Creation} and \textit{Direct Control}, generate and transmit the control commands from the operator to the vehicle. Reflecting the structure of the \ac{ros} packages in the code repository, the color-coded groups leverage modularity and scalability. For instance, a new control concept would be introduced as another \textit{Control} package.

\subsection{Bridge}
\label{ch:bridge}
The \textit{Bridge} contains the packages to create the interface with various vehicle systems. It is assumed that certain data feeds, such as an odometry feed and other vehicle feedback, are available from the teleoperated vehicle. If this is the case, the data are transmitted through the \textit{Network} to the operator. Also, the \textit{Bridge} handles launching of the camera and LiDAR sensor drivers, the data of which are processed and transmitted by the \textit{Perception} packages. Finally, dependent on the control mode, see~Sec.~\ref{ch:ctrl}, the \textit{Bridge} is also responsible for forwarding the respective commands from the operator to the actuation system of the vehicle.

\subsection{Perception}
The \textit{Perception} packages are responsible for processing, transmitting, and finally preparing the visualization of the vehicle sensor data for the operator.

\subsubsection{Video}
\label{ch:video}
As the primary source of information in \ac{tod}, video streams of the cameras on the vehicle are transmitted to the operator. In the system described, this is done using GStreamer~\cite{gstreamer}, a modular framework for different multimedia streaming applications. Due to its speed advantages, the H.264 codec~\cite{h264} is used to compress the videos. The video streaming sessions are established and controlled by the GStreamer RTSP Server Library~\cite{gstrtsp}. \\
The complete video streaming framework of the system offers a lot of flexibility. This includes adaptable parameters, such as the encoder bitrate, the video resolution scaling factor and cropping of the videos. The reconfiguration can be done either manually or automatically. The full functionality of the framework is described in~\cite{Schimpe2021}.

\subsubsection{Lidar}
The \textit{Lidar} package handles the transmission of the data from an array of LiDAR sensors on the vehicle to the operator. There, the data can be displayed in the \textit{Visual} package, see~Sec.~\ref{ch:visual}, or projected over the video, see~Sec.~\ref{ch:proj}. In addition, the \textit{Lidar} package also processes the data in certain ways for other \ac{ad} functions. For instance, object lists are generated from laser scans performing naive euclidean clustering. Also, a grid map, containing the occupancy of the vehicle surroundings, is constructed.

\subsubsection{Projection}
\label{ch:proj}
\begin{figure}[!t]
    \includegraphics[width=\linewidth]{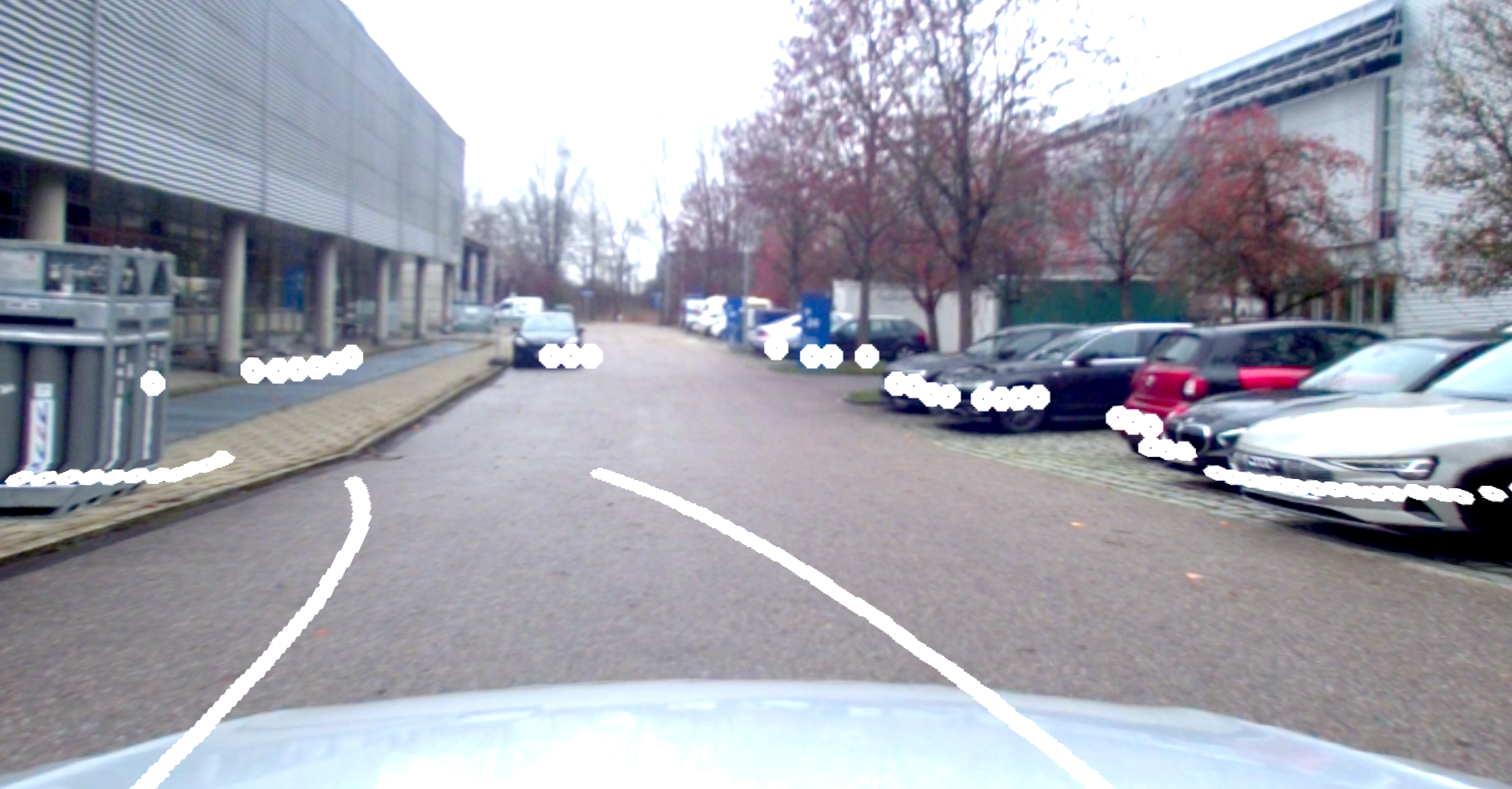}
    \caption{Projection of vehicle lane and laser scan points on video.}
    \label{fig:projection}
\end{figure}%
Based on vehicle feedback and sensor data, the \textit{Projection} package generates visual projections, which support the operator during teleoperation. For instance, the operator's anticipation of the future vehicle motion is improved through a projection of the vehicle lane. Assuming the current \ac{swa} value remains constant, the kinematic vehicle model equations are used to predict the locations of the left and right vehicle front edges. The obtained lanes are then superimposed on the video streams as a texture in the \textit{Visual} package, using the pinhole camera model and the OpenCV library~\cite{Bradski2000}. Fig.~\ref{fig:projection} shows a snapshot of the video that is displayed to the operator. In addition to the vehicle lane, the image also shows projections of the reflections from a 2D front laser scanner.

\subsection{Network}
A central part of the \ac{tod} system is communication via the network in order to connect vehicle and operator sides. In the system described, almost all the data is transmitted by the \textit{Network} package. The package contains templated sender-receiver pairs to transmit serialized \ac{ros} messages. In general, latency-critical data, such as control commands or LiDAR data, are transferred via~UDP. Other data, such as the system status message from the \textit{Manager} package, see~Sec.~\ref{ch:manager}, are transferred over TCP using \ac{mqtt}~\cite{IanCraggs}.

\subsection{Operator Interface}
The \textit{Operator Interface} consists of the packages that either display data to the operator or offer an interface that the operator can interact with. 

\subsubsection{Manager}
\label{ch:manager}
The \textit{Manager} package is used to establish and manage the status of the connection between the operator and the vehicle. A \ac{gui} with several functionalities is provided to the operator. Firstly, the IP addresses of the operator side and vehicle in question can be entered. Secondly, the actual teleoperation, i.e., transmission of control commands, can be started and stopped. Thirdly, the operator can switch between different input devices, see~Sec.~\ref{ch:inputDevs}, vehicle control modes, see~Sec.~\ref{ch:ctrl}, and video rate control modes, see~Sec.~\ref{ch:video}.

\subsubsection{Visual}
\label{ch:visual}
\newcommand{\rColW}{0.80} % right column width
\begin{table*}[!ht]
%\normalsize
\caption{Entities created in the visual scene and their properties.}
\centering
\begin{tabular}{| c | c |} \hline
	\textbf{Entity} & \begin{minipage}[c]{\rColW\textwidth} \centering \vspace*{1mm} \textbf{Description} \vspace*{1mm} \end{minipage} \\\hline

Scene Camera & \begin{minipage}[c]{\rColW\textwidth} \vspace*{1mm}
	Camera capturing the scene of the visualized entities. Follows the position of the vehicle model. \vspace*{1mm} \end{minipage} \\\hline

Coordinate Frame & \begin{minipage}[c]{\rColW\textwidth} \vspace*{1mm}
	Coordinate systems to describe relative positions of data, e.g., laser scans relative to the sensor. Used to transform positions into the coordinate system of the vehicle model to coherently render the complete scene. \vspace*{1mm} \end{minipage} \\\hline

Vehicle Model & \begin{minipage}[c]{\rColW\textwidth} \vspace*{1mm}
	The 3D model of the vehicle. Its position is continuously updated based on the odometry feed received from the vehicle. \vspace*{1mm} \end{minipage} \\\hline

Speedometer & \begin{minipage}[c]{\rColW\textwidth} \vspace*{1mm}
	An alpha-numeric display that shows the commanded and actual velocities of the vehicle as well as gear positions. \vspace*{1mm} \end{minipage} \\\hline

Video Canvas & \begin{minipage}[c]{\rColW\textwidth} \vspace*{1mm}
	A surface used for spherical or rectangular video data visualization. For a spherical projection, the underlying presumption is that the calibration of the sensor stack, both intrinsic and extrinsic, is provided. \vspace*{1mm} \end{minipage} \\\hline

Laser Scan & \begin{minipage}[c]{\rColW\textwidth} \vspace*{1mm}
	The reflections from obstacles, captured by a laser scan sensor. \vspace*{1mm} \end{minipage} \\\hline

Vehicle Lane & \begin{minipage}[c]{\rColW\textwidth} \vspace*{1mm}
	Projected motion of the vehicle, calculated based on the current steering wheel angle. Can also be used to display a different path, e.g., the vehicle motion planned by an automated driving function. \vspace*{1mm} \end{minipage} \\\hline

Top View & \begin{minipage}[c]{\rColW\textwidth} \vspace*{1mm}
	Rectangular display of the visual scene from above, captured by another scene camera. \vspace*{1mm} \end{minipage} \\\hline

\end{tabular}
\label{tab:entities}
\end{table*}%

\begin{figure}[!t]
    \includegraphics[width=\linewidth]{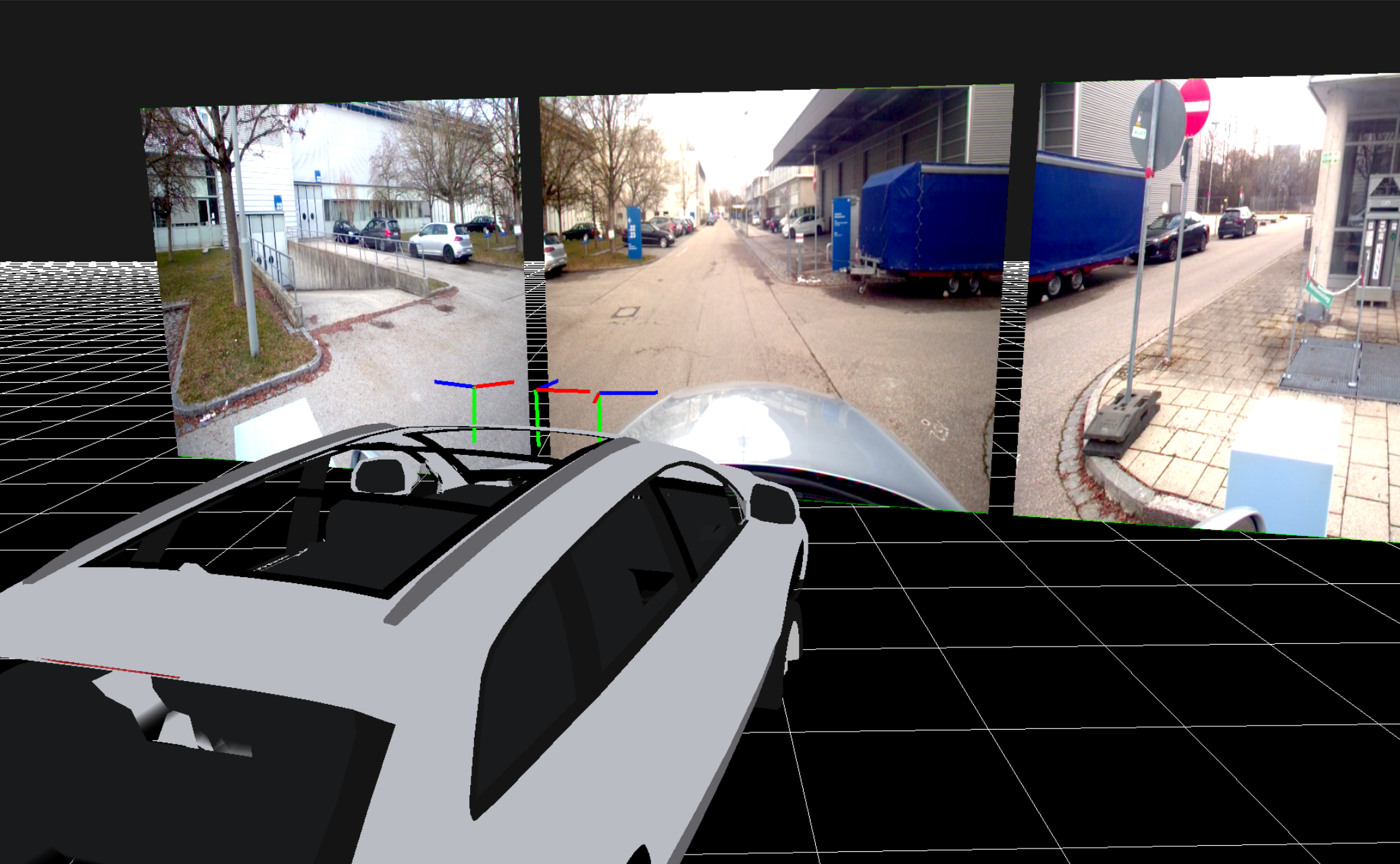}
    \caption{View of the vehicle model and videos from three front-mounted cameras rendered as rectangles in the \ac{hmi} of the \textit{Visual} package. In addition, RGB color-coded coordinate frames of the camera sensors are visible.}
    \label{fig:visVehView}
\end{figure}%

\begin{figure}[!t]
    \includegraphics[width=\linewidth]{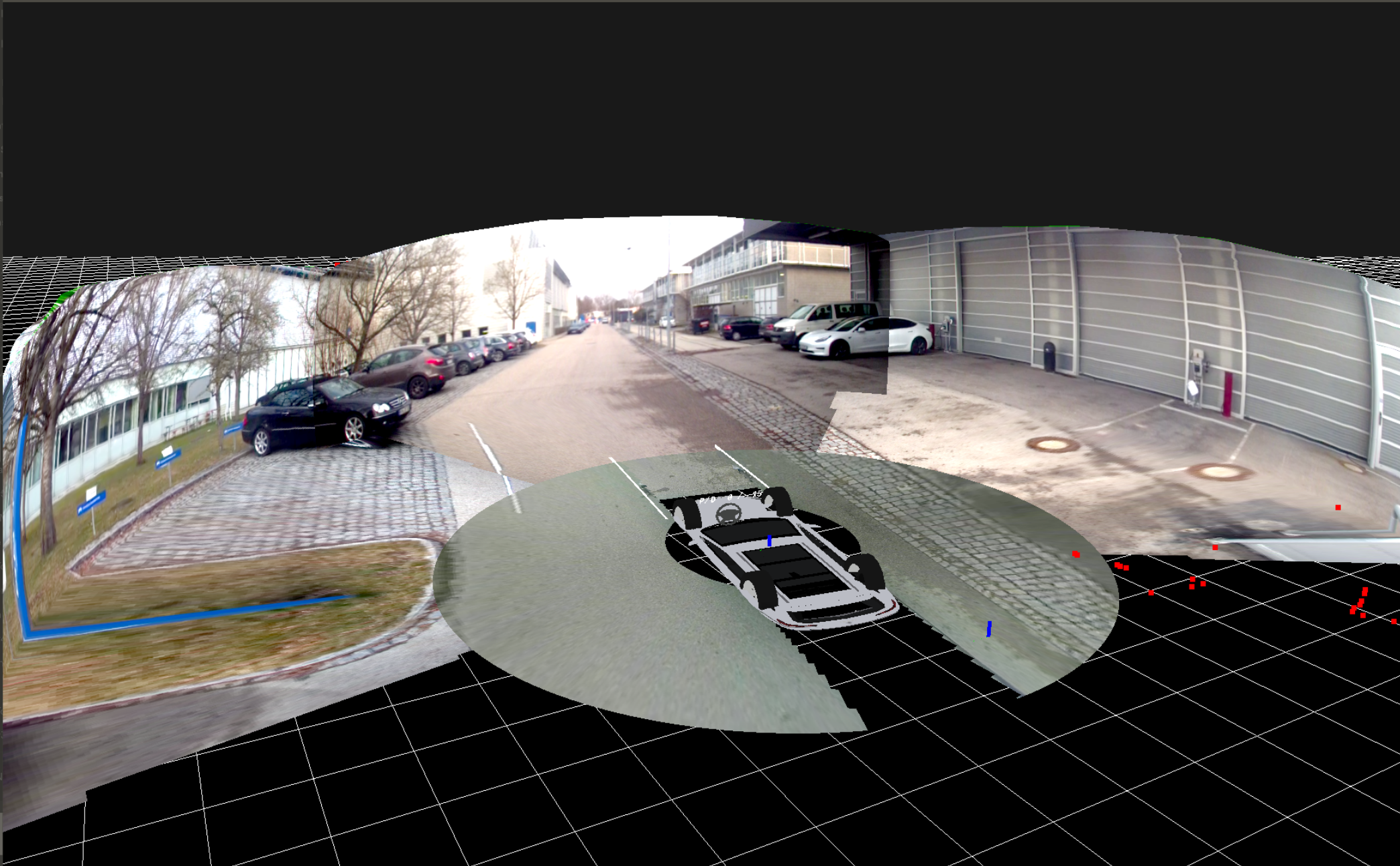}
    \caption{View of the display of videos from three front-mounted and three fisheye cameras on a spherical canvas in the \ac{hmi} of the \textit{Visual} package. In front of the vehicle, the lane of the projected vehicle motion is shown in white.}
    \label{fig:visSphView}
\end{figure}

The \textit{Visual} package provides a function-rich and flexible \ac{hmi} for the operator to perform the teleoperation. Adopting multiple concepts from~\cite{Georg2019b}, it displays received data from the \textit{Perception} and \textit{Bridge} packages in various ways. \\
A 3D world, comparable to the rviz package~\cite{rviz}, is constructed using the OpenGL API. Inspired by the open source game engine Hazel~\cite{HazelGithub}, the \textit{Visual} package uses the \ac{ecs} design pattern through the entt library~\cite{Caini}, based on the composition over inheritance principle. An overview of selected entities, i.e., the objects constructed and visualized in the \ac{hmi}, is given in Table~\ref{tab:entities}. Support for an \ac{hmd} has also been introduced. This provides an even more immersive experience during teleoperation. Snapshots of the \ac{hmi}, exhibiting a view of the vehicle model and giving an impression of the projection of video streams on either rectangles or a spherical canvas, are shown in Fig.~\ref{fig:visVehView} and Fig.~\ref{fig:visSphView}.\\
The aforementioned \ac{ecs} pattern has been shown to offer great flexibility. It leaves the package with great potential for future improvements and extensions. 

\subsubsection{Input Devices}
\label{ch:inputDevs}
The \textit{Input Devices} interface supports various input device hardware. For instance, the package is compatible with multiple USB devices. It also provides a virtual joystick that can be used without additional hardware. The outputs of this package are continuous axis values and discrete button states. In configuration mode, the allocation of axes and buttons to signals, such as the desired \ac{swa} or change in gear position, can be adapted for new input devices.

\subsection{Control}
\label{ch:ctrl}
Generally speaking, \textit{Control} packages generate, transmit and process control commands. Based on the output of the \textit{Input Devices}, the control commands are generated in the \textit{Command Creation} package and split into
\begin{itemize}
\item the primary control commands, which concern the lateral and longitudional motion of the vehicle,
\item and the secondary control commands, e.g., the gear position or indicators.
\end{itemize}
The control architecture has been developed with the objective of supporting a multitude of control modes. In \textit{Direct Control} mode, the primary control commands are directly transmitted to the vehicle for execution, i.e., the operator controls the vehicle at stabilization level. Other control modes, such as a shared control approach~\cite{Schimpe2020} or a concept to modify the perception of the vehicle \cite{Feiler2021} are also under development and will be integrated into the software stack. 

\section{\uppercase{Usability}}
\label{sec:usability}
The presented software stack has been developed and designed with the objective of making it applicable for a multitude of \acp{ugv}, and to support \ac{tod} research. The \ac{ros} framework is used to ease maintainability and modularity of the system. Hard-coding of parameters, such as actuation limits, sensor names or coordinate frame identifiers, is strictly avoided. Instead, these are specified in configuration files. Based on these, the system components construct and provide their functionality. For instance, the robot transform tree, video pipelines of the RTSP server, or sensor-receiver pairs for the transmission of laser scans are instantiated based on lists of transforms and sensors specified in the configuration files. In addition, to finally use the presented software stack to teleoperate an arbitrary \ac{ugv}, a \textit{Bridge} package, interfacing the vehicle-specifc sensors and actuators as shown in Fig.~\ref{fig:architecture}, needs to be provided. This package itself is assumed to provide a ROS interface that follows a certain straightforward topic naming convention. 

\section{\uppercase{Demonstrations}}
\label{sec:demo}
Flexilibity and usability, as discussed in the previous section, are demonstrated through the deployment of the software on a full-size passenger car, a~1:10-scale RC car, and a driving simulator. A video, showcasing the demonstrations, is available\footnote{\href{https://youtu.be/bQZLCOpOAQc}{https://youtu.be/bQZLCOpOAQc}}.
\subsection{Passenger Vehicle}
\label{sec:demo_q7}
\begin{figure}[!t]
    \includegraphics[width=\linewidth]{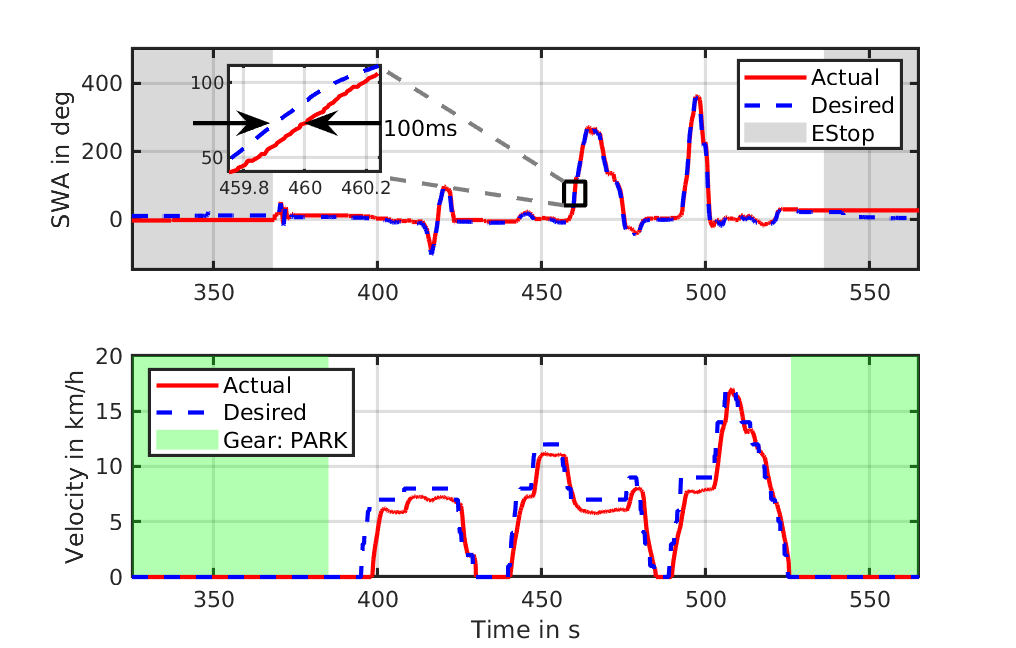}
    \caption{Steering wheel angle~(SWA) tracking~(top) and velocity tracking~(bottom) during teleoperation of the passenger vehicle. Actual values from the vehicle and desired values input by the operator are plotted against time. In addition, engagement of the emergency stop~(EStop) and the gear of the vehicle while in park position are shown. Enlargement of SWA plot at~$460$\,s exhibits actuation latency of approximately~$100$\,ms, as perceived by the operator.}
    \label{fig:q7track}
\end{figure}%
In this demonstration, a full-size passenger vehicle, an Audi Q7, is teleoperated. The sensor setup consists of three front-mounted, and one rear-mounted camera. Four fisheye cameras, mounted at the front and rear bumper, and below the left and right side mirrors, enable the operator to monitor the close surroundings of the vehicle. One~2D~laser scanner is also mounted on each of the front and rear bumpers. All sensors are connected via~USB~3.0 or Ethernet to the vehicle~PC, which is equipped with an Intel Xeon Gold~6130~\SI{2.10}{\giga\hertz}~16 core processor and~\SI{32}{\giga\byte} of~RAM. Commands are written to, and feedback read from the vehicle~CAN bus through a dSpace Autobox, also connected to the vehicle~PC via Ethernet. The end-to-end delay, the so-called \ac{g2g} latency~\cite{Bachhuber2016}, of a~\SI{40}{\hertz}, 520p video feed, transmitted over a wired connection and displayed to the operator on a gaming monitor, operating at~\SI{144}{\hertz}, is approximately~\SI{104}{\milli\second}. A thorough assessment and comparisons of the latency for different configurations within the same system are provided in~\cite{Georg2020}. \\
The described vehicle is teleoperated on private roads for approximately~\SI{100}{\second}. A safety driver is located in the vehicle, who releases the emergency stop and is ready to take control of the vehicle at all times. The teleoperation is started and stopped at a standstill. The driving course consists of a narrow lane change, defined by foam cubes, two left turns and two stop lines. The performance of the tracking of the~\ac{swa} and the velocity are shown in Fig.~\ref{fig:q7track}. Upon release of the emergency stop, it can be seen that the actual~\ac{swa} accurately follows the commands from the operator. In the zoomed-in section at around~\SI{460}{\second}, the experienced actuation latency of around~\SI{100}{\milli\second} becomes apparent. As the signals are logged on the operator side, this actuation latency includes~\textit{(1)}~the transmission of the operator's command to the vehicle and therein to the dSpace Autobox,~\textit{(2)}~the actual latency of the steering actuator,~\textit{(3)}~reading of the actual~\ac{swa} value from the vehicle~CAN bus, and~\textit{(4)}~the transmission of this signal back to the operator. The velocity, desired by the operator, is also tracked reliably. After shifting gear from park to drive, the vehicle drives at moderate speeds up to~\SI{17}{\kilo\meter\per\hour}, coming to a stop twice during driving and finally at the end of the course. Minor stationary tracking errors can be observed. However, these do not have a negative effect on the performance of the operator during teleoperation.
\subsection{RC Car}
\begin{figure}[!t]
    \includegraphics[width=\linewidth]{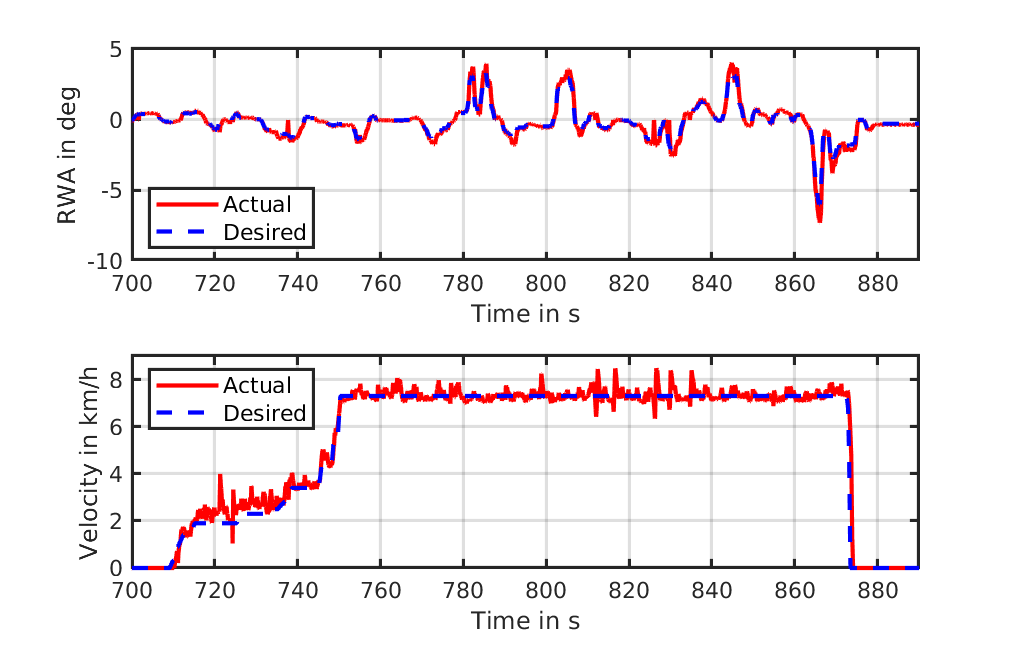}
    \caption{Road wheel angle~(RWA) tracking~(top) and velocity tracking~(bottom) during teleoperation of the RC car. Actual vehicle values and values desired by the operator are plotted against time. The actual value signals were smoothed to remove noise, and improve clarity.}
    \label{fig:rctrack}
\end{figure}%
The presented software stack has also been applied to teleoperate an F1TENTH, 1:10-scale RC~car~\cite{OKelly2019}. The only sensor used is a front-facing stereo camera. On board, the software runs on an NVIDIA Jetson TX2. At~\SI{241}{\milli\second}, the \ac{g2g} latency for a \SI{15}{\hertz}, 520p video feed of the RC~car is significantly larger when compared to the passenger vehicle. However, it is expected that optimizations such as a higher framerate will reduce the \ac{g2g} latency of the RC~car. \\
The RC~car is teleoperated on the same private roads for approximately~\SI{190}{\second}. Fig.~\ref{fig:rctrack} depicts the tracking performance of the \ac{rwa} and the velocity, commanded by the operator. The actual \ac{rwa} follows the desired \ac{rwa} reliably. While lower velocities are tracked erratically, the performance beyond \SI{3}{\kilo\meter\per\hour} is consistent. 
\subsection{Driving Simulator}
\begin{figure*}[!t]
	\includegraphics[width=\linewidth]{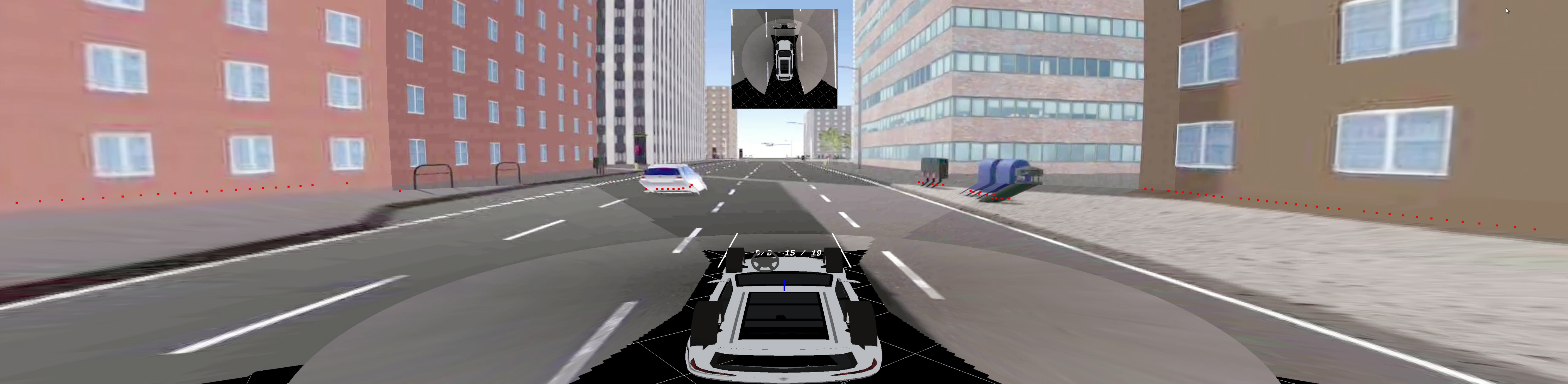}
	\caption{Operator's view while teleoperating the driving simulator.}
	\label{fig:lgsvl_wide}
\end{figure*}%
The SVL (formerly LGSVL) driving simulator~\cite{Rong2020} is handled as another vehicle system within the software stack. The SVL bridge package provides conversion nodes for the interface required by the actual simulator bridge~(V2020.06). A widescreen operator view is shown in Fig.~\ref{fig:lgsvl_wide}. The simulation was run on the vehicle PC, described in Sec.~\ref{sec:demo_q7}. The sensor configuration of the simulated vehicle was specified so that it was similar to that of the passenger vehicle. 

\section{\uppercase{Conclusion}}
\label{sec:conclusion}
This paper has presented an open source software stack for \ac{tod}. As teleoperation of \acp{ugv} and especially road vehicles is receiving increasing attention in research and industry, the publication of the software is aimed at supporting research in this field. During development, the emphasis was on modularity and scalability to facilitate deployment of the software on different vehicles with only minor overheads. In this paper, this was demonstrated with three systems, namely a full-size passenger vehicle, a 1:10-scale RC~car and a driving simulator. While further technical aspects, such as innovative control concepts, are to be developed and validated in the future, the software will also be used for research involving human-machine interaction studies.

\section*{\uppercase{Acknowledgements}}
Andreas Schimpe, Johannes Feiler, Simon Hoffmann and Domagoj Majstorovi\'c as the first authors, were the initiators and main developers of the presented work. The demonstration with the RC~car was supported by Florian Sauerbeck. Special thanks are also due to Jean-Michael Georg for his continuous advice on software development. Frank Diermeyer made essential contributions to the conception of the research project and revised the paper critically for important intellectual content. He gave final approval for the version to be published and agrees to all aspects of the work. As a guarantor, he accepts responsibility for the overall integrity of the paper. The research was partially funded by the European Union (EU) under RIA grant No.~825050, the Federal Ministry of Education and Research of Germany~(BMBF) within the project UNICARagil (FKZ~16EMO0288), the Central Innovation Program~(ZIM) under grant No.~ZF4648101MS8, and through basic research funds from the Institute for Automotive Technology.

\bibliographystyle{IEEEtranBST/IEEEtran}
\bibliography{IEEEtranBST/IEEEabrv,Literatur}

\begin{thebibliography}{10}
\providecommand{\url}[1]{#1}
\csname url@rmstyle\endcsname
\providecommand{\newblock}{\relax}
\providecommand{\bibinfo}[2]{#2}
\providecommand\BIBentrySTDinterwordspacing{\spaceskip=0pt\relax}
\providecommand\BIBentryALTinterwordstretchfactor{4}
\providecommand\BIBentryALTinterwordspacing{\spaceskip=\fontdimen2\font plus
\BIBentryALTinterwordstretchfactor\fontdimen3\font minus
  \fontdimen4\font\relax}
\providecommand\BIBforeignlanguage[2]{{%
\expandafter\ifx\csname l@#1\endcsname\relax
\typeout{** WARNING: IEEEtran.bst: No hyphenation pattern has been}%
\typeout{** loaded for the language `#1'. Using the pattern for}%
\typeout{** the default language instead.}%
\else
\language=\csname l@#1\endcsname
\fi
#2}}

\bibitem{Davis2010}
J.~Davis, C.~Smyth, and K.~McDowell, ``{The Effects of Time Lag on Driving
  Performance and a Possible Mitigation},'' \emph{IEEE Transactions on
  Robotics}, vol.~26, no.~3, pp. 590--593, 2010.

\bibitem{Tang2014e}
T.~Tang, P.~Vetter, S.~Finkl, K.~Figel, and M.~Lienkamp, ``{Teleoperated road
  vehicles - The "Free Corridor" as a safety strategy approach},''
  \emph{Applied Mechanics and Materials}, vol. 490-491, pp. 1399--1409, 2014.

\bibitem{Schimpe2021}
A.~Schimpe, S.~Hoffmann, and F.~Diermeyer, ``{Adaptive Video Configuration and
  Bitrate Allocation for Teleoperated Vehicles},'' in \emph{Proc. of Workshop
  for Road Vehicle Teleoperation (RVT) at 2021 IEEE Intelligent Vehicles
  Symposium (IV21)}, 2021.

\bibitem{Mutzenich2021}
C.~Mutzenich, S.~Durant, S.~Helman, and P.~Dalton, ``{Updating our
  understanding of situation awareness in relation to remote operators of
  autonomous vehicles},'' \emph{Cognitive Research: Principles and
  Implications}, vol.~6, no.~1, 2021.

\bibitem{Bensoussan1997}
S.~Bensoussan and M.~Parent, ``{Computer-aided teleoperation of an urban
  vehicle},'' in \emph{Proc. of 8th International Conference on Advanced
  Robotics (ICAR)}.\hskip 1em plus 0.5em minus 0.4em\relax IEEE, 1997, pp.
  787--792.

\bibitem{Gnatzig2013a}
S.~Gnatzig, F.~Chucholowski, T.~Tang, and M.~Lienkamp, ``{A system design for
  teleoperated road vehicles},'' in \emph{Proc. of 10th International
  Conference on Informatics in Control, Automation and Robotics (ICINCO)},
  vol.~2, Reykjav{\'{i}}k, 2013, pp. 231--238.

\bibitem{Bodell2016}
O.~Bodell and E.~Gulliksson, ``{Teleoperation of Autonomous Vehicle With 360°
  Camera Feedback},'' Master's Thesis, Chalmers University of Technology, 2016.

\bibitem{Shen2016}
X.~Shen, Z.~J. Chong, S.~Pendleton, G.~M.~J. Fu, B.~Qin, E.~Frazzoli, and M.~H.
  Ang, ``{Teleoperation of on-road vehicles via immersive telepresence using
  off-the-shelf components},'' \emph{Advances in Intelligent Systems and
  Computing}, vol. 302, no. September 2016, pp. 1419--1433, 2016.

\bibitem{Georg2019b}
J.-M. Georg and F.~Diermeyer, ``{An adaptable and immersive real time interface
  for resolving system limitations of automated vehicles with teleoperation},''
  in \emph{Proc. of IEEE International Conference on Systems, Man and
  Cybernetics (SMC)}.\hskip 1em plus 0.5em minus 0.4em\relax IEEE, 2019, pp.
  2659--2664.

\bibitem{Hofbauer2020}
M.~Hofbauer, C.~B. Kuhn, G.~Petrovic, and E.~Steinbach, ``{TELECARLA: An Open
  Source Extension of the CARLA Simulator for Teleoperated Driving Research
  Using Off-the-Shelf Components},'' in \emph{Proc. of IEEE Intelligent
  Vehicles Symposium (IV)}.\hskip 1em plus 0.5em minus 0.4em\relax IEEE, 2020,
  pp. 335--340.

\bibitem{Dosovitskiy2017}
A.~Dosovitskiy, G.~Ros, F.~Codevilla, A.~L{\'{o}}pez, and V.~Koltun, ``{CARLA:
  An open urban driving simulator},'' in \emph{Proc. of 1st Annual Conference
  on Robot Learning (CoRL)}, 2017, pp. 1--16.

\bibitem{Kato2018}
S.~Kato, S.~Tokunaga, Y.~Maruyama, S.~Maeda, M.~Hirabayashi, Y.~Kitsukawa,
  A.~Monrroy, T.~Ando, Y.~Fujii, and T.~Azumi, ``{Autoware on Board: Enabling
  Autonomous Vehicles with Embedded Systems},'' in \emph{Proc. of 9th ACM/IEEE
  International Conference on Cyber-Physical Systems, (ICCPS)}, 2018, pp.
  287--296.

\bibitem{Baidu}
\BIBentryALTinterwordspacing
Baidu, ``{Apollo auto}.'' [Online]. Available: \url{http://apollo.auto/}
\BIBentrySTDinterwordspacing

\bibitem{Quigley2009}
\BIBentryALTinterwordspacing
M.~Quigley, B.~Gerkey, K.~Conley, J.~Faust, T.~Foote, J.~Leibs, E.~Berger,
  R.~Wheeler, and A.~Ng, ``{ROS: an open-source Robot Operating System},'' in
  \emph{ICRA Workshop on Open Source Software}, 2009, pp. 1--6. [Online].
  Available:
  \url{http://www.willowgarage.com/papers/ros-open-source-robot-operating-system}
\BIBentrySTDinterwordspacing

\bibitem{gstreamer}
\BIBentryALTinterwordspacing
{GStreamer Team}, ``gstreamer - open source multimedia framework.'' [Online].
  Available: \url{https://gstreamer.freedesktop.org/}
\BIBentrySTDinterwordspacing

\bibitem{h264}
\BIBentryALTinterwordspacing
{VideoLAN Organization}, ``x264 - a free open-source h.264 encoder.'' [Online].
  Available: \url{http://www.videolan.org/developers/x264.html}
\BIBentrySTDinterwordspacing

\bibitem{gstrtsp}
\BIBentryALTinterwordspacing
{GStreamer Team}, ``{GStreamer RTSP Server}.'' [Online]. Available:
  \url{http://gstreamer.freedesktop.org/modules/gst-rtsp-server.html}
\BIBentrySTDinterwordspacing

\bibitem{Bradski2000}
G.~Bradski, ``{The OpenCV Library},'' \emph{Dr. Dobb's Journal of Software
  Tools}, 2000.

\bibitem{IanCraggs}
\BIBentryALTinterwordspacing
I.~Craggs, ``{The Paho C library}.'' [Online]. Available:
  \url{https://github.com/eclipse/paho.mqtt.cpp}
\BIBentrySTDinterwordspacing

\bibitem{rviz}
\BIBentryALTinterwordspacing
ROS-Team, ``rviz.'' [Online]. Available: \url{http://wiki.ros.org/rviz}
\BIBentrySTDinterwordspacing

\bibitem{HazelGithub}
\BIBentryALTinterwordspacing
Y.~Chernikov, ``{Hazel Engine}.'' [Online]. Available:
  \url{https://github.com/TheCherno/Hazel}
\BIBentrySTDinterwordspacing

\bibitem{Caini}
\BIBentryALTinterwordspacing
M.~Caini, ``{EnTT}.'' [Online]. Available:
  \url{https://skypjack.github.io/entt/}
\BIBentrySTDinterwordspacing

\bibitem{Schimpe2020}
A.~Schimpe and F.~Diermeyer, ``{Steer with Me: A Predictive, Potential
  Field-Based Control Approach for Semi-Autonomous, Teleoperated Road
  Vehicles},'' in \emph{Proc. of 23rd International Conference on Intelligent
  Transportation Systems (ITSC)}.\hskip 1em plus 0.5em minus 0.4em\relax IEEE,
  2020.

\bibitem{Feiler2021}
J.~Feiler and F.~Diermeyer, ``{The Perception Modification Concept to Free the
  Path of An Automated Vehicle Remotely},'' in \emph{Proc. of 7th International
  Conference on Vehicle Technology and Intelligent Transport Systems (VEHITS
  2021)}.\hskip 1em plus 0.5em minus 0.4em\relax SciTePress, 2021, pp.
  405--412.

\bibitem{Bachhuber2016}
C.~Bachhuber and E.~Steinbach, ``{A System for High Precitision Glass-to-Glass
  Delay Measurements in Video Communication},'' in \emph{Proc. of IEEE
  International Conference on Image Processing (ICIP)}.\hskip 1em plus 0.5em
  minus 0.4em\relax Phoenix, AZ, USA: IEEE, 2016, pp. 8--12.

\bibitem{Georg2020}
J.-M. Georg, J.~Feiler, S.~Hoffmann, and F.~Diermeyer, ``{Sensor and Actuator
  Latency during Teleoperation of Automated Vehicles},'' in \emph{Proc. of IEEE
  Intelligent Vehicles Symposium (IV)}, 2020, pp. 760--766.

\bibitem{OKelly2019}
\BIBentryALTinterwordspacing
M.~O'Kelly, H.~Abbas, J.~Harkins, C.~Kao, Y.~V. Pant, R.~Mangharam, V.~S. Babu,
  D.~Agarwal, M.~Behl, P.~Burgio, and M.~Bertogna, ``{F1/10: An open-source
  autonomous cyber-physical platform},'' 2019. [Online]. Available:
  \url{https://arxiv.org/abs/1901.08567}
\BIBentrySTDinterwordspacing

\bibitem{Rong2020}
G.~Rong, B.~H. Shin, H.~Tabatabaee, Q.~Lu, S.~Lemke, M.~Mo{\v{z}}eiko,
  E.~Boise, G.~Uhm, M.~Gerow, S.~Mehta, E.~Agafonov, T.~H. Kim, E.~Sterner,
  K.~Ushiroda, M.~Reyes, D.~Zelenkovsky, and S.~Kim, ``{LGSVL Simulator: A High
  Fidelity Simulator for Autonomous Driving},'' in \emph{Proc. of 23rd
  International Conference on Intelligent Transportation Systems,
  (ITSC)}.\hskip 1em plus 0.5em minus 0.4em\relax IEEE, 2020.

\end{thebibliography}

\end{document}